\documentclass[12pt]{article}
\usepackage[a4paper,top=1.5cm,bottom=1.5cm,left=2.5cm,right=2cm,marginparwidth=1.5cm]{geometry}
\usepackage[utf8]{inputenc}
\usepackage{graphicx}
\usepackage{amsmath}
\usepackage{caption}
\usepackage{subcaption}
\usepackage{tabularx}

\usepackage{dingbat}
\usepackage{etoolbox,lineno, xcolor}
\usepackage{lipsum}

\usepackage{authblk}
\usepackage{hyperref}
\usepackage{xurl}
\usepackage{biblatex}
\begin{document}
\title{\textbf{MIDAL: A Dataset of Math Image Descriptions for Accessible Learning}}
\author[1]{Rebeka Popek \footnote{Primary Author.\\
\raisebox{-5pt}{\includegraphics[width=0.04\linewidth]{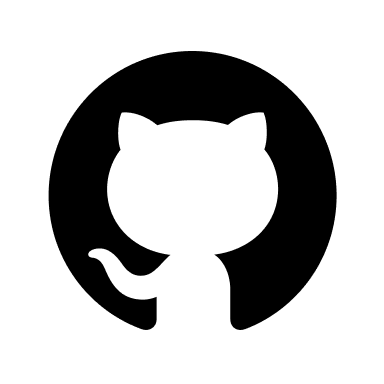}}~Code: \href{https://github.com/Rebeka-P/MIDAL-Dataset}{\nolinkurl{Rebeka-P/MIDAL-Dataset}}\\
\raisebox{-7pt}{\includegraphics[width=0.04\linewidth]{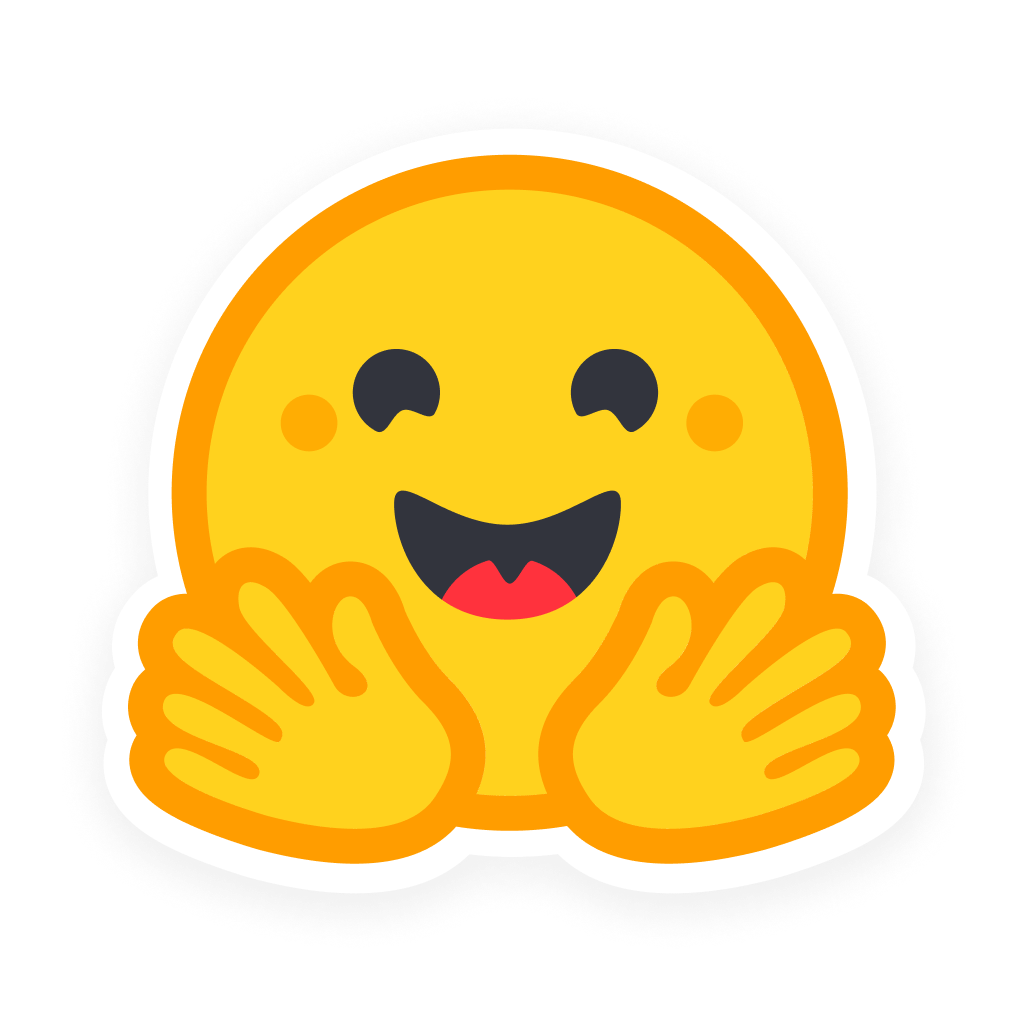}}~Dataset: \href{https://huggingface.co/datasets/rpopek/MIDAL}{\nolinkurl{rpopek/MIDAL}}}}
\author[2]{Vaghawan Ojha} 
\author[3]{Young Hwan You}
\affil[1]{\href{rpopek@iu.edu}{\texttt{rpopek@iu.edu}}, \href{rebekapopek@gmail.com}{\texttt{rebekapopek@gmail.com}}}
\affil[2]{\href{vaghawan.ojha@ekbana.net}{\texttt{vaghawan.ojha@ekbana.net}}}
\affil[3]{\href{youy@iu.edu}{\texttt{youy@iu.edu}}}
\affil[1, 3]{School of Natural Science and Mathematics, Indiana University East}
\affil[2] {Department of AI, E.K. Solutions Pvt. Ltd.}
\date{}
\maketitle

\section*{Abstract}
Many open educational resources are lacking in accessibility, especially in-depth image descriptions. In subjects like Science and Mathematics, however, it can be particularly difficult to write image descriptions since there can be many complicated expressions and names depending upon the course level. To help fill that gap in a small way, we introduce Math Image Descriptions for Accessible Learning (MIDAL), a math image-description dataset of 2,020 mathematical images spanning multiple educational levels, to aid in training vision language models to create image descriptions following accessibility best practices. We hope MIDAL is a valuable resource in enhancing the conversation and innovation regarding accessibility of STEM content in higher education. This dataset is however not just limited in math description generation but can also be used to fine-tune language models that can have improved mathematical reasoning and answers.
\section{Introduction}
The United States’ Department of Justice issued a final rule in April 2024, updating the regulations for Title II of the Americans with Disabilities Act (ADA) \cite{ada_web_rule_2024} to require state and local governments to modify all public-facing web and mobile content to be accessible up to the standard Web Content Accessibility Guidelines (WCAG) Version 2.1, Level AA \cite{w3c_wcag21_2025}. If met, these guidelines ensure a color contrast ratio of 4.5:1 in text and images of text, the ability to resize onscreen text without loss of functionality, the inclusion of headings, and alternative text (alt text) for images, among other requirements that are out of the scope of this paper \cite{w3c_wcag21_2025}. All qualifying departments and institutions must be in compliance by April 24, 2026, but "state and local entities with a population of 50,000 or more" have been granted an extension until April 26, 2027 \cite{ada_web_rule_2024}.
Science, technology, engineering, and math (STEM) subjects are quite visual (e.g., graphs, charts, diagrams, complex math equations), which presents a challenge to blind or low-vision students. Limited access to accessible materials magnifies difficulties already faced, including poorer math performance \cite{gatto2024accessiblemath}. According to a survey of 355 open access textbooks, 80.34\% of textbooks with images failed to provide alternative text \cite{azadbakht2021notopen}. Alt text is a type of image description that is embedded in an image's metadata, but has a character limit \cite{perkins}. Long image descriptions are used for complex graphics that cannot be accurately described within the character limit, and is often displayed elsewhere on the page \cite{perkins}. 
We present Math Image Descriptions for Accessible Learning (MIDAL), a dataset with 2,020 mathematical images, long descriptions, figure captions, and paragraph context, to help address the lack of sufficient alt text and image descriptions in textbooks and online instructional materials for mathematics. The content of the dataset covers elementary, high school, and college-level mathematics. This dataset is intended for use in image descriptions of mathematical images and helping train large language models on mathematical figure reasoning. It is licensed under CC BY-NC-SA 4.0.

\section{MIDAL Composition}

The MIDAL dataset was composed through a three-stage data collection and curation workflow designed to ensure both accessibility quality and usable research structure. First, we identified and selected freely available online mathematics textbooks that already provided high-quality, human-written image descriptions (e.g., for figures, diagrams, plots, and other visual elements commonly used in math instruction). Second, from the selected sources, we systematically extracted the relevant images along with their surrounding textual context (such as nearby paragraphs, captions, section headings, and referenced equations) and paired these materials with the corresponding image descriptions to preserve the intended meaning of each visual. Third, we performed dataset labeling and post-processing, which included verifying and cleaning the extracted pairs, standardizing formats, organizing metadata, and applying controlled manipulations where needed (e.g., normalization, filtering, deduplication, or restructuring) to produce a consistent, well-annotated corpus suitable for downstream modeling and evaluation. The complete procedure is described in detail in the sections that follow.

\subsection{Image Description and Sources}
Recently, there has been a surge of free online textbooks for students and teachers. Unfortunately, many of the free textbooks are often not accessible for all students. There are numerous high-quality mathematics textbooks that are only available as PDF files, which are not inherently accessible to students who rely on screen readers, Braille displays, or other assistive technology~\cite{USC_math_acc}. This narrowed the search to mathematics textbooks hosted on websites or as HTML files. Although there was still a significant number of textbooks and resources available in this fashion, far less were available with a Creative Commons copyright, and even fewer with alternative text paired with their images.
The search for free, accessible math images was conducted for over a month. Websites such as Pressbooks, LibreTexts, University of Minnesota’s Open Textbook Library, and Open Educational Resources (OER) Commons were used to facilitate this search. Any resource that was open access and displayed on a website was considered, and the HTML was analyzed for quality alternative text.
Content was downloaded from the following establishments:
\begin{itemize}
    \item Indiana University~\cite{iu_oer}
    \item Rice University’s OpenStax~\cite{openstax_about}
    \item Portland Community College~\cite{pcc_oer}
    \item Pennsylvania State University~\cite{pennstate_oer}
    \item Virginia Military Institute~\cite{vmi_oer_policy}
    \item Stephen F. Austin State University~\cite{sfa_oer}
    \item BCcampus Open Education~\cite{bccampus_opened}
    \item University of Lethbridge~\cite{uleth_oer}
    \item British Columbia Institute of Technology~\cite{bcit_open}
    \item Maricopa Community College~\cite{maricopa_oer}
    \item University of Northern Colorado~\cite{unco_oer}
    \item Texas A\&M University~\cite{tamu_opened}
    \item University of Sheffield~\cite{sheffield_oer}
    \item Alamo Colleges District~\cite{alamoopen_oer}
    \item Pierce College~\cite{pierce_oer}
    \item Rhode Island College~\cite{ric_oer}
    \item Open Oregon Educational Resources~\cite{openoregon_oer}
\end{itemize}

The complete list of e-texts can be found in the appendix.
There were two primary means of collection: an automated Python script to scrape website data and manual collection. We created a Python script, utilizing the open-source Python library Playwright \cite{playwrightFastReliable} for web browser automation. The data collected included all images, context in from the surrounding text, the figure captions, the page URL, and the book title. The second method of collection, manual collection, was used for websites with inconsistent image layouts or websites containing many images that were not necessary or needed for our purposes. Any images that were SVG files were also converted to PNG or JPG files with a white background using Inkscape’s~\cite{Inkscape} command line interface.

\subsection{Data Labeling and Manipulation}
Label Studio \cite{label-studio}, an open-source data annotation tool, was used to consolidate and adjust the images and their descriptions. To begin, fifty images were loaded into Label Studio. Using the information collected from the Python script and through manual collection from the website itself, the data was inputted physically into each of the following fields:
\begin{itemize}
    \item Image description
    \item Image context
    \item Image caption
    \item Book
    \item Webpage URL
\end{itemize}
The following two images are screenshots from the data annotation editor. \newpage
\begin{figure}
    \centering
    \includegraphics[width=0.5\linewidth, alt={Parabolic function with a hole and points labeled.}]{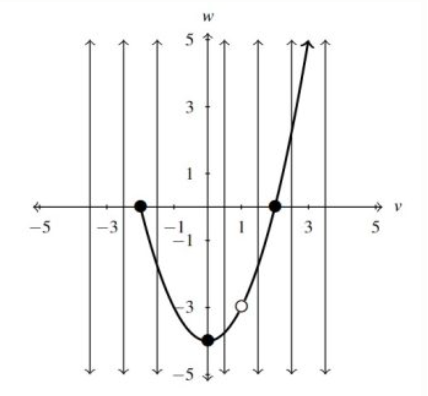}
    \caption{Source: Carl Stitz and Jeff Zeager, \textit{Functions, Trigonometry, and Systems of Equations} (2024), used under CC BY-NC-SA 4.0 International.}
    \label{fig:figure-sample}
\end{figure}

\begin{figure}[!h]
    \centering
    \includegraphics[width=0.65\linewidth, alt={Screenshot of Label Sudio annotation for the previous image, with text input boxes completed for the following sections: Describe the image, Context, Figure caption.}]{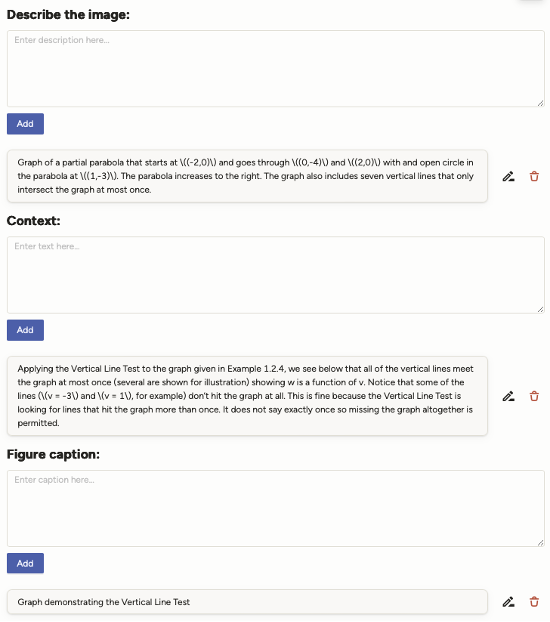}
    \caption{Screenshot of data annotation in Label Studio for the previous image, Fig \ref{fig:figure-sample}, 2025.
}
    \label{fig:data-annotation}
\end{figure}

Data labeling started in February 2025 and ended in June 2025. OpenAI’s ChatGPT \cite{chatgptChatGPT} and Microsoft’s Copilot \cite{microsoftMicrosoftCopilot} were used to speed up transcribing paragraphs with math content into LaTeX. To have greater control over the consistency of the descriptions (the “reference text”), every description was manually reviewed, evaluated, and corrected before being added to the dataset. 
After 2,020 data entries were created, the data needed to be cleaned due to all text fields being human generated. JavaScript was used to modify the data JSON file (i.e., adjusting field names, consistent textbook names, adjusting any missing values) and Python was used to concatenate and upload the modified JSON file to Hugging Face as a Hugging Face dataset. Although consistency was strived for throughout the creation of MIDAL, the data was created by hand over the course of four months; there may be some errors.

\subsection{Copyrights}
All images and content compiled in MIDAL were obtained from online sources that explicitly provide content under a Creative Commons copyright license. Each item was reviewed to ensure compliance with the license and terms of use, unless specifically granted permission in writing. In the dataset, every image has the name of the website or textbook the content was taken from and the URL of the webpage the content appears. To be in compliance with the diverse Creative Commons licenses, MIDAL is licensed under Creative Commons Attribution-NonCommercial-ShareAlike 4.0 International (CC BY-NC-SA 4.0). It should be noted that OpenStax does not allow for large language models to train on the textbooks without prior permission, which the authors received before creating the dataset.

\subsection{A Closer Look at MIDAL}
For proper credit, all data entries have an abbreviated book title attributed to it. Those with the value “rebeka” were created by the authors, so there is not a source URL for the entries. The majority (97.82\%) of entries have “context” attributed to them, which is any surrounding long text that gives a student background for an image. In contrast, captions are shorter texts, either included in the textbook as a captioned image, or a brief label created by the authors. There are captions present for 35.64\% of the entries.
In the next sections, we quantitatively describe MIDAL through mathematical subjects, mathematical education level, image size, and image blurriness.

\subsubsection*{Mathematical Content Diversity}
In Table \ref{tab:math-subjects}, we show the approximate mathematical subject diversity across the curated dataset. 

  \begin{table}[!h]
\centering
    \begin{tabular}{|l|c|c|}
    \hline
        \textbf{Mathematical Subject} & \textbf{Number} & \textbf{Percentage (\%)}\\
        \hline
        Abstract Algebra & 27 & 1.34\%\\
        \hline
        Algebra & 747 & 36.98\%\\
        \hline
        Business Math & 4 & 0.20\%\\
        \hline
        Calculus & 316 & 15.64\%\\
        \hline
        Computational Biology & 16 & 0.79\%\\
        \hline
        Discrete Math & 44 & 2.18\%\\
        \hline
        Fundamental & 25 & 1.24\%\\
        \hline
        Geometry & 584 & 28.91\%\\
        \hline
        Precalculus & 253 & 12.52\%\\
        \hline
        Statistics & 4 & 0.20\%\\
        \hline
    \end{tabular}
      \caption{Distribution of mathematical figures per subject area.}
    \label{tab:math-subjects}
    \end{table}

\begin{figure}
    \centering
    \includegraphics[width=\linewidth, alt={Horizontal bar plot titled Images per Mathematical Subject. The following are subjects and their number of images in the dataset. Stastics: 4. Business math: 4. Computational biology: 16. Fundamental: 25. Abstract algebra: 27. Discrete math: 44. Precalculus: 253. Calculus: 316. Geometry: 584. Algebra: 747.}]{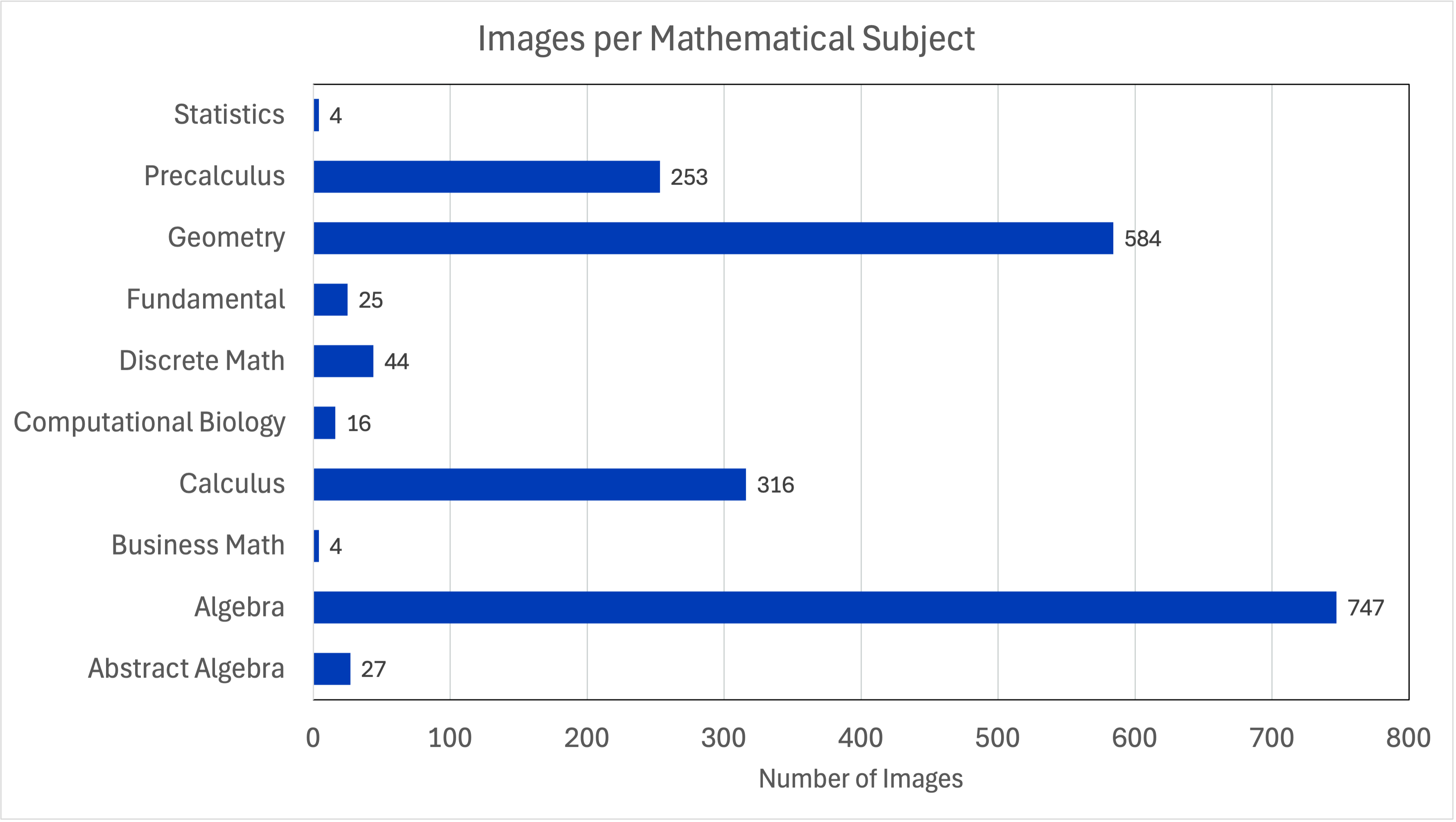}
    \caption{Images per subject}
    \label{fig:dist-histro}
\end{figure}

\begin{figure}
    \centering
    \includegraphics[width=0.7\linewidth, alt={Pie chart titled Images per Mathematical Education Level. College is 34\% of the total. High school is 65\% of the total. Elementary is 1\% of the total.}]{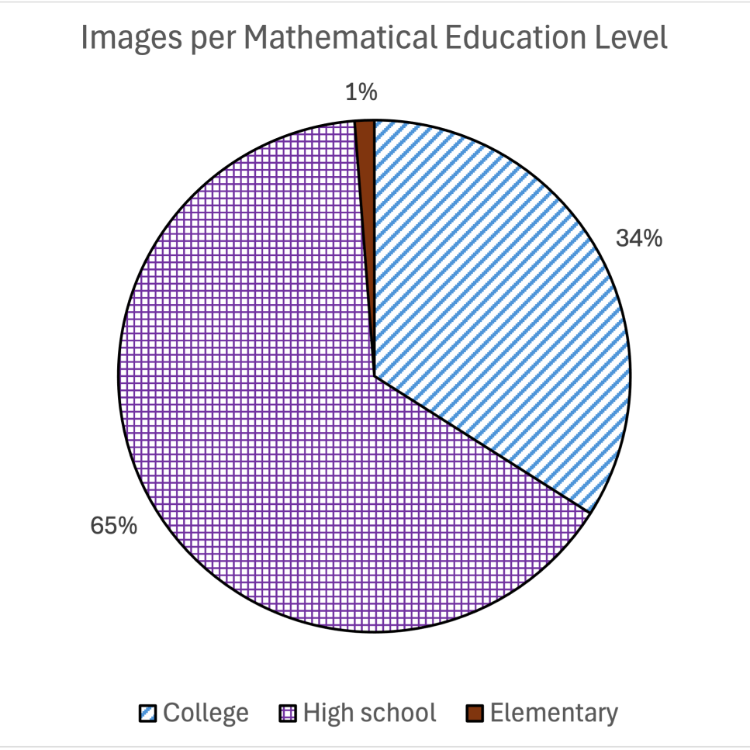}
    \caption{Images per education level}
    \label{fig:dist-pie}
\end{figure}

Additionally, majority of the images were at a high school level of mathematics, as seen in Figures \ref{fig:dist-histro} and \ref{fig:dist-pie}.

\subsubsection{Image Quality}
\begin{figure}[!h]
    \centering
    \includegraphics[width=0.9\linewidth, alt={Frequency Histogram of Total Image Pixel Resolutions with horizontal axis titled Total Pixel Resolution and the vertical axis titled Number of Images. See surrounding text for description.}]{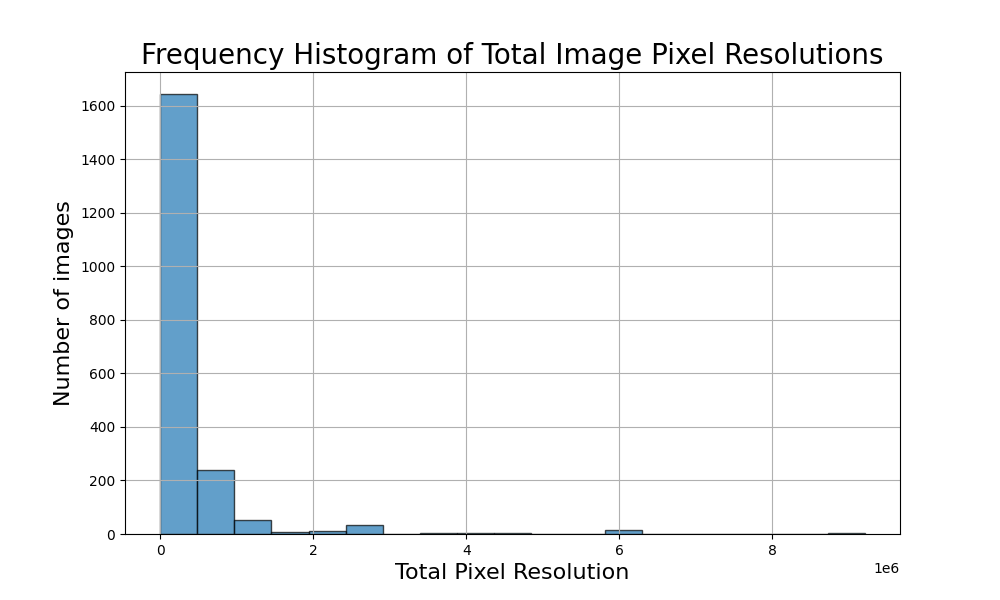}
    \caption{Pixel Resolutions}
    \label{fig:pixel-resolution}
\end{figure}

In MIDAL, the image resolutions are heavily skewed toward smaller pixel resolutions, as seen in Figure \ref{fig:pixel-resolution}. More than 1,600 images have a pixel resolution of 500k or smaller, which is about 79\% of the dataset. Few images are above 2 million pixels in resolution.

\begin{figure}[!h]
    \centering
    \includegraphics[width=0.8\linewidth, alt={Scatterplot titled Image Size vs Laplacian Variance with horizontal axis titled Image Size (Total Pixels) and vertical axis titled Laplacian Variance. There is no pattern but majority of the points are clustered between 0 to 2 million pixels and 0 to 10,000 laplacian variance.}]{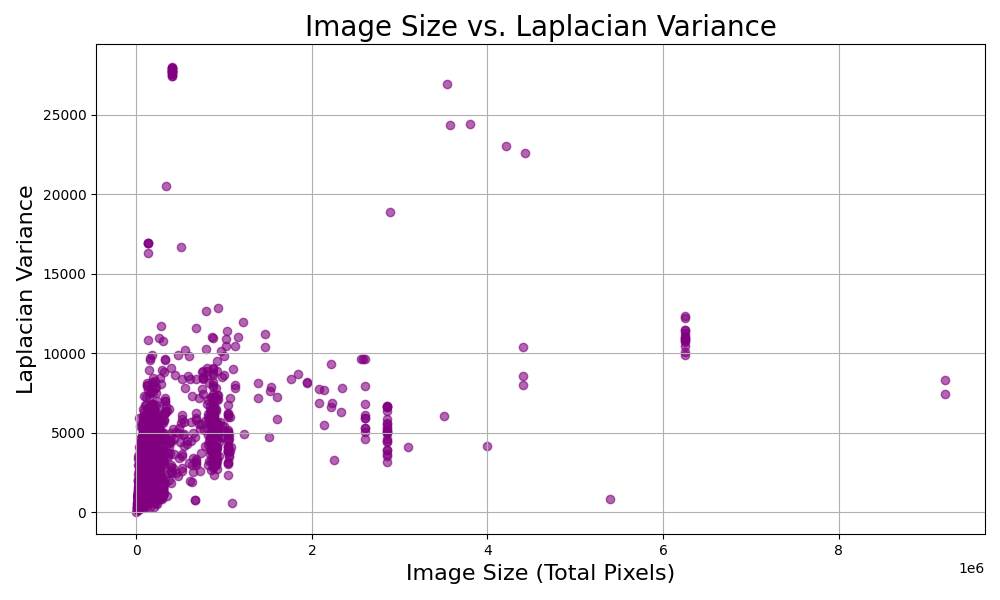}
    \caption{Normalized image size Laplacian}
    \label{fig:normalized}
\end{figure}

Normalized Laplacian variance from OpenCV \cite{opencv_library} (all images resized to 256 by 256 pixels) was used to quantify the sharpness of images in the dataset. Figure \ref{fig:normalized} shows a scatterplot between total image resolution in pixels and the normalized Laplacian variance. There is no obvious correlation between image resolution and the normalized Laplacian variance.

\begin{figure}[!h]
    \centering
    \includegraphics[width=0.8\linewidth, alt={Histogram of Laplacian Variance Scores with horizontal axis titled Laplacian Variance and vertical axis titled Number of Images. See description in surrounding text.}]{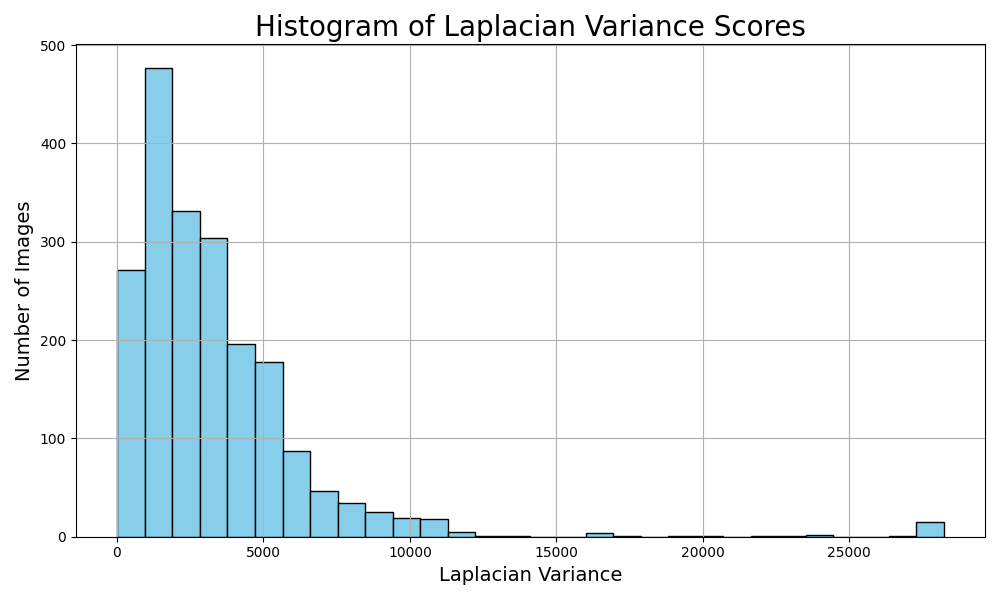}
    \caption{Overall normalized Laplacian variance scores}
    \label{fig:normalized-fig}
\end{figure}

Figure \ref{fig:normalized-fig} shows a histogram with a right-skew of the Laplacian variance. Most images have a low normalized Laplacian variance score, less than 5000.

\begin{figure}[!h]
    \centering
    \includegraphics[width=\linewidth, alt={Top 5 Most Blurry Images and their scores. The lowest score here is 8.33. The highest score here is 179.18.}]{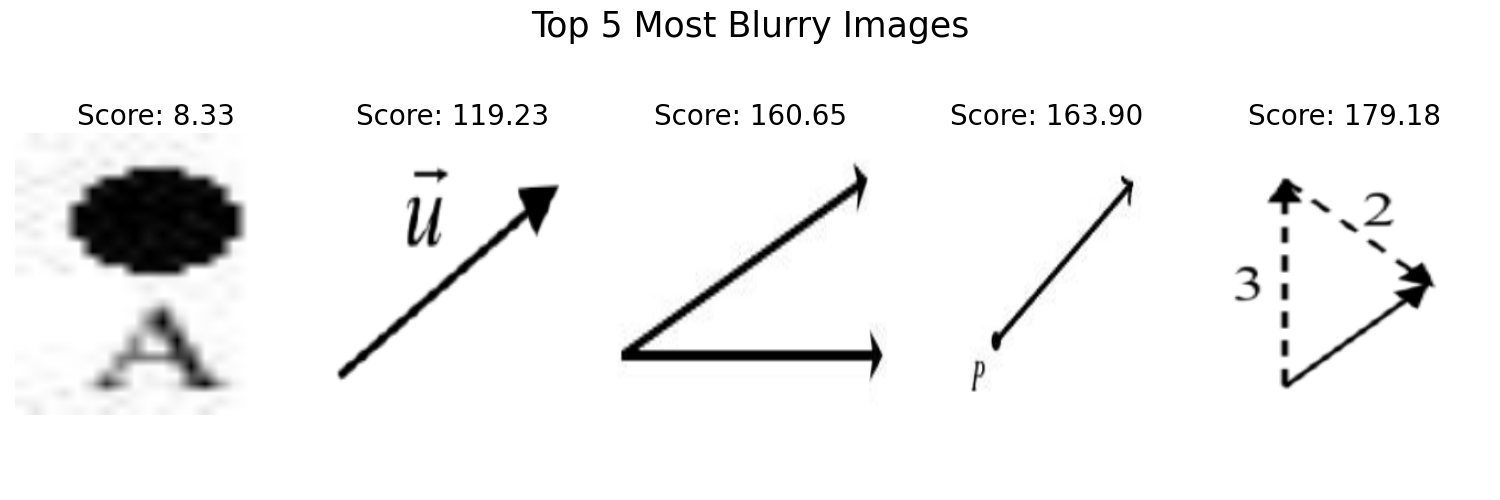}
    \caption{Images in dataset with lowest scores (most blurry)}
    \label{fig:blurry}
\end{figure}

\begin{figure}[!h]
    \centering
    \includegraphics[width=\linewidth, alt={Top 5 Least Blurry Images and their scores. The lowest score here is 28222.08. The highest score here is 27980.65.}]{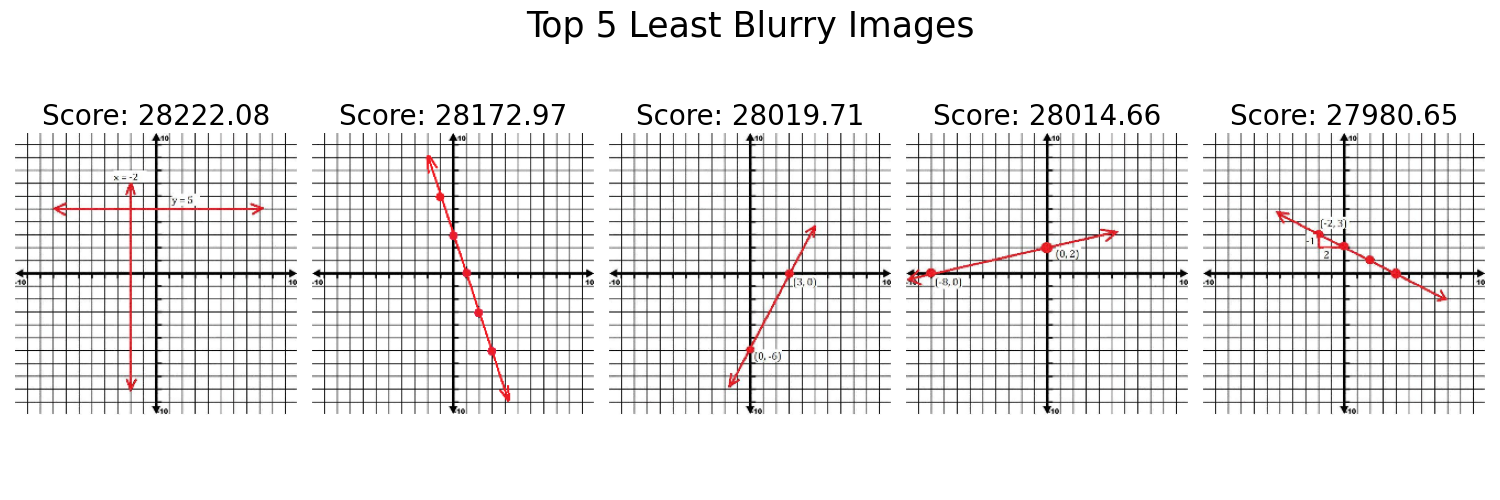}
    \caption{Images in dataset with highest scores (least blurry)}
    \label{fig:higher-blurry}
\end{figure}

Figures \ref{fig:blurry} and \ref{fig:higher-blurry} show the five most blurry and five least blurry images (respectively) using the normalized Laplacian variance score.

\clearpage

\section{Related Work}
There are many large image-caption datasets, including the COYO-700 Image-Text Pair Dataset (2022) \cite{kakaobrain2022coyo}, Microsoft COCO (2015) \cite{lin2015microsoftcococommonobjects}, VizWiz-Captions dataset (2020) \cite{Gurari_2019_CVPR}, and PixelProse (2024) \cite{singla2024pixelsproselargedataset}. While necessary, these datasets are for general captioning tasks. There is a large gap in the current data regarding academic images and their respective descriptions. Recently, there has been an increase in mathematically related datasets like DynaMath (2025) \cite{zou2025dynamathdynamicvisualbenchmark} and DrawEduMath (2025) \cite{baral2025drawedumathevaluatingvisionlanguage} aim to enhance VLM problem-solving and reasoning.
Our dataset combines mathematical images, their contexts, and image descriptions that follow NCAM \cite{ncam} guidelines for conciseness, accuracy, and clarity. A related dataset created for a different purpose is the MM-Math-Align dataset \cite{sun2025hardnegativecontrastivelearning}, which aims to improve geometric reasoning tasks through “negative contrastive learning”.
\section{Limitations and Conclusion}
The MIDAL dataset is comprised of images from online open educational resources. Although we strived to incorporate mathematical diversity in the content, we were limited by time and what images were already accessible. This resulted in a small dataset. As stated earlier, more than 80\% of open access textbooks do not contain alt text at all. Additionally, the dataset is only in English and lacks hand-drawn images. Despite these limitations, MIDAL is the first dataset of its kind to combine mathematical images with their descriptions based in documented accessibility best practices. 

\subsection*{Acknowledgements}
This research was supported in part by Lilly Endowment, Inc., through its support for the Indiana University Pervasive Technology Institute, Quartz high throughput computing cluster. This project was financially supported by the Indiana University East Summer Research Scholars (SUMRS) Program. Author Vaghawan, would like to thank E.K. Solutions (Ekbana) for its support on time and resources provided to him during the execution of this project.

\clearpage
\printbibliography

\clearpage
\section*{Books Used in the Dataset}

\begin{center}
    \begin{tabular}{|>{\raggedright\arraybackslash}m{0.30\textwidth}|%
    >{\raggedright\arraybackslash}m{0.30\textwidth}|%
    >{\raggedright\arraybackslash}m{0.30\textwidth}|}
        \hline
        \textbf{Author(s)} & \textbf{Title} & \textbf{Book URL}\\
        \hline
        Abramson, J. and Falduto, V. and Gross, R. and Lippman, D. and Rasmussen, M. and Norwood, R. and Belloit, N. and Whipple, H. and Magnier, J. M. and Fernandez, C. & Algebra and Trigonometry, 2nd ed & \url{https://openstax.org/details/books/algebra-and-trigonometry-2e}\\
        \hline
        Arendt, K. and De La Nuez, M. and Girard, L. & Adult Literacy Fundamental Mathematics: Book 4, 2nd ed & \url{https://opentextbc.ca/alfm4/}\\
        \hline
        Arthur, M. M. L. and Clark, R. & Social Data Analysis & \url{https://pressbooks.ric.edu/socialdataanalysis/}\\
        \hline
        Avilez, A. and Ceinaturaga, S. and Levine, T. D. & College Mathematics -- MAT14X, 3rd ed & \url{https://open.maricopa.edu/collegemathematicstextbookmat14x2ndedition/}\\
        \hline
        Best, A. & Introducing Mathematical Biology & \url{https://sheffield.pressbooks.pub/introducingmathematicalbiology/}\\
        \hline
        Chase, M. & Technical Mathematics, 2nd ed & \url{https://openoregon.pressbooks.pub/techmath2e/}\\
        \hline
        Coffelt, V. & Functions, Trigonometry, and Systems of Equations & \url{https://odp.library.tamu.edu/math150/}\\
        \hline
        Flinn, C. and Overgaard, M. & Math for Trades: Volume 2 & \url{https://opentextbc.ca/mathfortrades2/}\\
        \hline
        Hartman, G. & APEX Calculus & \url{https://opentext.uleth.ca/apex-calculus/apex-calculus.html}\\
        \hline
        Jordan, A. and Cary, A. and Kouzes, R. and Leavitt, S. and Lee, C. and Yao, C. and Youtz, R. & Open Resources for Community College Algebra, 3rd ed & \url{https://spot.pcc.edu/math/orcca/ed3/html/orcca.html}\\
        \hline
        Judson, T. W. and Beezer, R. A. & Abstract Algebra: Theory and Applications & \url{https://judsonbooks.org/aata-files/aata-html/aata-toc.html}\\
        \hline
        Kellman, C. and Major, L. and Mallory, D. and Gruen, F. and Goldlist, A. & Business Mathematics & \url{https://pressbooks.bccampus.ca/businessmathematics/}\\
        \hline
        Levin, O. & Discrete Mathematics: An Open Introduction, 4th ed & \url{https://discrete.openmathbooks.org/dmoi4/dmoi4.html}\\
        \hline
        Lin, K. & Matrices & \url{https://psu.pb.unizin.org/matricesversiontwo/}\\
        \hline
    \end{tabular}
\end{center}

\begin{center}
    \begin{tabular}{|>{\raggedright\arraybackslash}m{0.30\textwidth}|%
    >{\raggedright\arraybackslash}m{0.30\textwidth}|%
    >{\raggedright\arraybackslash}m{0.30\textwidth}|}
        \hline
        \textbf{Author(s)} & \textbf{Title} & \textbf{Book URL}\\
        \hline
        Lippman, D. and Rasmussen, M. & Precalculus: An Investigation of Functions & \url{https://math.libretexts.org/Bookshelves/Precalculus/Book%3A_Precalculus__An_Investigation_of_Functions_(Lippman_and_Rasmussen)}\\
        \hline
        Liu, L. & Math Mastery Manual & \url{https://acd.pressbooks.pub/math-mastery-manual/}\\
        \hline
        Pennsylvania State University & MATH 140 - Calculus and Analytic Geometry 1 | Calculus and Analytic Geometry & \url{https://sites.psu.edu/math140jaj/}\\
        \hline
        Murphy, S. & The Art of Polynomial Interpolation & \url{https://psu.pb.unizin.org/polynomialinterpretation/}\\
        \hline
        Indiana University & OnRamp Math Site & \url{https://expand.iu.edu/browse/iu-online/courses/onramp-math-site}\\
        \hline
        Overgaard, M. and Flinn, C. & Math for Trades: Volume 3 & \url{https://opentextbc.ca/mathfortrades3/}\\
        \hline
        Strang, G. and Herman, E. and Bila, N. V. and Boyd, S. J. and Smith, D. and Terry, E. A. and Torain, D. and Messer, K. R. and Mulzet, A. K. and Radulovich, W. and Rutter, E. M. and McCune, D. and Merriweather, M. and Lakey, J. and Levandosky, J. and Falduto, V. and Abbott, C. and Debnath, J. & Calculus: Volume 3 & \url{https://openstax.org/details/books/calculus-volume-3}\\
        \hline
        Strang, G. and Herman, E. and Radulovich, W. and Rutter, E. M. and Smith, D. and Messer, K. R. and Mulzet, A. K. and Bila, N. V. and Boyd, S. J. and Debnath, J. and Falduto, V. and Terry, E. A. & Calculus: Volume 2 & \url{https://openstax.org/details/books/calculus-volume-2}\\
        \hline
        Tagami, W. and Girard, L. & Adult Literacy Fundamental Mathematics: Book 2, 2nd ed & \url{https://opentextbc.ca/alfm2/}\\
        \hline
        Tagami, W. and Girard, L. & Adult Literacy Fundamental Mathematics: Book 3, 2nd ed & \url{https://opentextbc.ca/alfm3/}\\
        \hline
    \end{tabular}
\end{center}

\end{document}